\documentclass{article}
\usepackage{spconf,amsmath,epsfig}
\usepackage{hyperref}

\title{PREDICTION OF ANNUAL SNOW ACCUMULATION USING A RECURRENT GRAPH CONVOLUTIONAL APPROACH}
\name{Benjamin Zalatan$^1$, Maryam Rahnemoonfar$^{1,2,*}$ \thanks{$^*$Corresponding author (maryam@lehigh.edu).}}

\address{
$^1$Department of Computer Science and Engineering, Lehigh University, PA, USA\\
$^2$Department of Civil and Environmental Engineering, Lehigh University, PA, USA
}
\begin{document}
%
\maketitle
\begin{abstract}
The precise tracking and prediction of polar ice layers can unveil historic trends in snow accumulation. In recent years, airborne radar sensors, such as the Snow Radar, have been shown to be able to measure these internal ice layers over large areas with a fine vertical resolution. In our previous work, we found that temporal graph convolutional networks perform reasonably well in predicting future snow accumulation when given temporal graphs containing deep ice layer thickness. In this work, we experiment with a graph attention network-based model and used it to predict more annual snow accumulation data points with fewer input data points on a larger dataset. We found that these large changes only very slightly negatively impacted performance.
\end{abstract}
\begin{keywords}
deep learning, graph neural networks, ice thickness, remote sensing
\end{keywords}
\section{Introduction}
\label{sec:intro}

The accurate tracking and forecasting the internal ice layers of polar ice sheets is important for reducing the uncertainties in current climate model predictions and future sea level rise. They allow for the precise calculation of snow mass balance, extrapolation of ice age from direct measurements of the subsurface, and inferring otherwise difficult to observe ice dynamic processes.

Measurements of ice layers are traditionally collected via ice cores and shallow pits. However, spatial sparsity, access difficulty, and high cost make it challenging to capture catchment-wide accumulation rates. Attempts to interpolate these in-situ measurements introduce additional uncertainties due to the high variability in local accumulation rate.

Airborne measurements using nadir-looking radar sensors has become a popular method of mapping ice topography and monitoring accumulation rates with a broad spatial coverage. They also have the ability to reveal and capture the structure of isochronous ice layers beneath the ice sheet surface. Subsurface ice layers that appear in the resulting echograms can unveil historic annual and multi-annual snow accumulation as well as help in quantifying the impact of atmospheric warming on polar ice caps and project its contribution to future sea level rise.


We focus on the Snow Radar \cite{snow-radar} dataset collected by the Center for Remote Sensing of Ice Sheets (CReSIS) as part of NASA's Operation IceBridge. The Snow Radar operates from 2-8 GHz and is able to track deep layers of ice with a high resolution over wide areas of an ice sheet. The sensor produces a two-dimensional grayscale profile of historic snow accumulation over consecutive years, where the horizontal axis represents the along-track direction, and the vertical axis represents layer depth. Pixel brightness is directly proportional to the strength of the returning signal. Pixels representing surface layers are generally brighter and more well-defined due to high reflectance and snow density variation, while pixels representing deeper layers are generally darker and noisier due to increased density and a lower return-signal strength. In our experiments, we use radar data from selected Snow Radar flights over Greenland in the year 2012. In many areas, each ice layer represents an annual isochrone \cite{koenig2016annual}. As such, we may refer to specific ice layers by their corresponding year.

In a previous paper \cite{zalatan2023recurrent}, we proposed a model designed to predict future snow accumulation when given the thicknesses of deep ice layers. In this paper, adjust this model and its surrounding methods in the following ways:

\begin{enumerate}
    \item We use five deep ice layers to predict ten shallow ice layers, rather than ten deep ice layers to predict five shallow ice layers.
    \item We use a substantially larger dataset (1254 images rather than 568) that includes a validation set.
    \item We increase the complexity of the model by adding an additional fully-connected layer, using surface elevation as a node feature, and implementing weight decay.
    \item We use a graph attention network (GAT) \cite{veličković2018graph} rather than simple graph convolution.
\end{enumerate}




\begin{figure*}
    \centerline
    {
        \includegraphics[width=\textwidth]{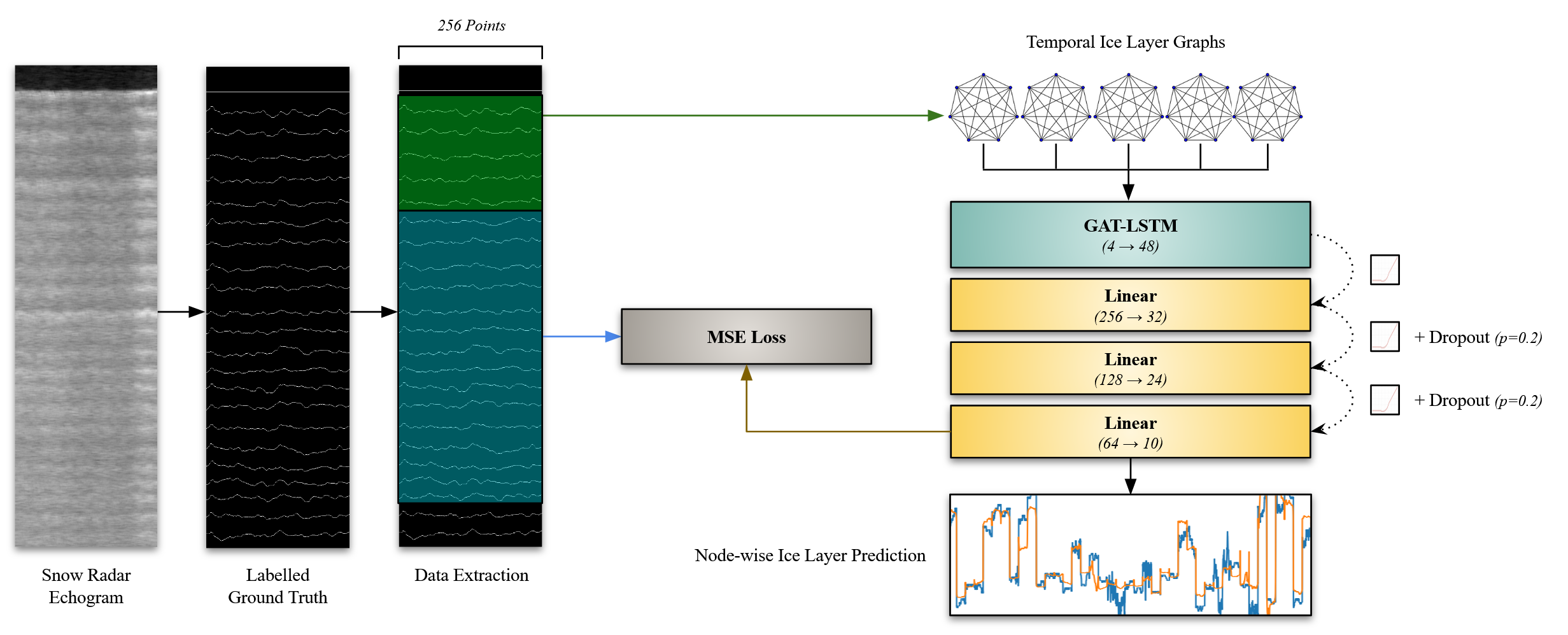}
    }
    \caption{Architecture of the proposed model.}
    \label{Fig1:arch}
\end{figure*}

\section{Related Work}
\label{sec:relwork}

Automated techniques, such as \cite{Rahnemoonfar-level-set, Rahnemoonfar-charged-particles, Rahnemoonfar_AIRadar}, have been developed for detecting the surface and bottom layers of echograms using radar depth sounder sensors. Deep learning has also been applied on Snow Radar data for tracking the internal layers of polar ice sheets \cite{rahnemoonfar2021deep, varshney2020deep, yari2019smart, yari2020multi}. \cite{yari2019smart} used a multi-scale contour-detection CNN to segment the different internal ice layers found in Snow Radar images. The authors of this work also experimented with pretraining on the Berkeley Segmentation Dataset and Benchmark (BSDS) dataset \cite{arbelaez2010contour} and found that this was not very effective due to the large amount of noise in Snow Radar images. In \cite{rahnemoonfar2021deep}, the authors trained a multi-scale neural network on synthetic Snow Radar images for more robust training. A multi-scale network was also used in \cite{yari2020multi}, where the authors trained a model on images from the year 2012 and fine tuned it by training on a small number of images from other years. \cite{varshney2020deep} found that using pyramid pooling modules, a type of multi-scale architecture, helps in learning the spatio-contextual distribution of pixels for a certain ice layer. The authors also found that denoising the images prior to the CNN layer improved both the accuracy and F-score. These works have all used multi-scale networks in order to segment the ice layers within Snow Radar images, but have commonly noted that the high amount of noise present in Snow Radar images is an issue that must be addressed in the future. While these works have attempted to track and segment the annual ice layers present within grayscale Snow Radar echograms, none have yet attempted to use spatiotemporal patterns of the ice layers to predict future snow accumulation given the thicknesses of deep ice layers. In our previous work \cite{zalatan2023recurrent} presented at the IEEE Radar Conference, we evaluated how a GCN-LSTM model would perform in predicting future snow accumulation when given deep ice layer thickness in the form of temporal graphs. We found that this model structure performed noticeably better than models using non-temporal GCN and non-geometric LSTM.

\section{Dataset}
\label{sec:dataset}

\begin{table*}
    \centering
    \caption{Results from the non-temporal, non-geometric, and proposed models on the ten predicted annual ice layer thicknesses from 2003 to 2012. Results are shown as the mean $\pm$ standard deviation of the RMSE over five trials (in pixels).}
    \begin{tabular}{ | c | c | c | c | c |  } 
        \hline
        & LSTM & GCN & GAT-LSTM  \\ 
        \hline
        Total RMSE & $6.914 \pm 0.945$ & $5.489 \pm 0.153$ & $4.768 \pm 0.372$ \\ 
        \hline
    \end{tabular}
    \label{table:OverallResults}
\end{table*}

In this study, we used Snow Radar data captured as part of NASA's Operation Ice Bridge mission \cite{snow-radar}. CReSIS has made this data publicly available on their website (\url{https://data.cresis.ku.edu/}). The data gathered during each flight was processed and separated into a series of grayscale echogram images, each with a width of $256$ pixels and a height ranging between $1200$ and $1700$ pixels. Each pixel in a column corresponds to approximately $4$ cm of ice, and each radar image has an along-track footprint of 14.5m. The value of each pixel is proportional to the relative received power of the returning radar signal at that depth. Accompanying each image was a $256 \times 3$ matrix specifying the geographic latitude, longitude, and surface elevation associated with each vertical column of pixels. In order to gather ground-truth thickness data, the images were manually labelled in a binary format where white pixels represented the tops of each firn layer, and all other pixels were black.

In order to capture a sufficient amount of data, only radar images containing a minimum of $15$ ice layers were used: ten shallow layers for ground truths, and five deeper layers for use as node features. Ten shallow layers and five deep layers were chosen in order to maximize the number of usable images while also maintaining a reasonable amount of data per image. This limitation reduced the total number of usable images from $3147$ down to $1254$. Five different training, validation, and testing sets were generated by taking five random permutations of all usable images and splitting them at a ratio of 3:1:1. Each training set contained $752$ images, each validation set contained $251$ images, and each testing set contained $251$ images.

\section{Methods}
\label{sec:methods}

Traditional CNNs use a matrix of learnable weights, often referred to as a kernel or filter, as a sliding window across pixels in an input image. This allows a model to automatically extract image features that would otherwise need to be identified and inputted manually. Graph convolutional networks apply similar logic to graphs, but rather than using a sliding window of weights across an image, GCN performs weighted-average convolution on each node's neighborhood to extract features that reflect the structure of a graph. The size of the neighborhood on which convolution takes place is dictated by the number of GCN layers present in the model (i.e. $K$ GCN layers result in $K$-hop convolution). In a sense, graph convolutional networks are a generalized form of traditional CNN's that enable variable degree and weighted adjacency.

Recurrent neural networks (RNNs) are able to process a sequence of data points as input, rather than a single static data point, and learn the temporal relationships between them. Many traditional RNN structures had issues with vanishing and exploding gradients on long input sequences. Long short-term memory (LSTM) attempts to mitigate those issues by implementing gated memory cells that guarantee constant error flow \cite{lstmpaper}. Applying LSTM to a graph neural network allows for a model to learn not only the relationships between nodes in a graph, but also how those relationships change (or persist) over time.

Graph attention networks implicitly consider edge weights in a graph as learned parameters, rather than static values. This allows for complex relationships between nodes to be automatically extracted and represented in the graph's adjacency matrix. In the context of ice layer prediction, it could theoretically allow edge weights to represent more significant relationships than simple geographic distance between nodes.

We use a GAT-LSTM layer with $48$ output channels, that lead into three fully-connected layers of $32$, $24$, and $10$ output channels respectively. The output of the final layer is the predicted thickness (in pixels) of the ten shallow ice layers immediately beneath the surface, representing annual snow accumulation from 2003 to 2012. Between each layer is the Hardswish activation function \cite{hardswishpaper}, chosen due to its superior performance when compared to ReLU and its derivatives \cite{swishbetter}. Between the fully-connected layers is Dropout \cite{dropoutpaper} with $p=0.2$. We use the Adam optimizer \cite{adampaper} with a weight decay of $0.0001$ and dynamic learning rate that begins at $0.01$ and halves every $125$ epochs. We train for $500$ epochs using mean-squared error loss. The architecture of this model is shown visually in Figure \ref{Fig1:arch}.

Each ground truth echogram is converted into five ``feature'' graphs, each consisting of $256$ nodes. Each graph corresponds to a single ice layer for each year from 1998 to 2002. Each node represents a vertical column of pixels in the radar echogram image and has four features: two for the latitude and longitude that correspond to that column, one for the surface elevation at that point, and one for the thickness of the corresponding year's ice layer within that column (in pixels). All graphs are fully connected and undirected. All edges are inversely weighted by the geographic distance between node locations using the haversine formula. For a weighted adjacency matrix $A$:

{\small
    \begin{equation*}
        A_{i, j} = \frac{1}{2\arcsin\bigg(\text{hav}(\phi_j - \phi_i) + \cos(\phi_i)\cos(\phi_j)\text{ hav}(\lambda_j - \lambda_i)\bigg)}
    \end{equation*}
}
where
{\small
\begin{equation*}
    \text{hav}(\theta) = \sin^2 \bigg(\frac{\theta}{2} \bigg)
\end{equation*}
}
$A_{i, j}$ represents the weight of the edge between nodes $i$ and $j$. $\phi$ and $\lambda$ represent the latitude and longitude features of a node, respectively. Node features of all graphs are collectively normalized using z-score normalization. Weights in the adjacency matrices of all graphs are collectively normalized using min-max normalization.

\section{Results}
\label{sec:results}

In order to verify that the temporal and geometric aspects of the model continue to serve to its benefit, we compared its performance with equivalent models that utilize a non-temporal GCN as well as a non-geometric LSTM. For the GCN model, all hyperparameters remain the same, but the GAT-LSTM layer sequence is replaced by a single GCN layer, and all temporal graphs are consolidated into a single static graph. For the LSTM model, all hyperparameters remain the same, but an LSTM layer is used, rather than a GAT-LSTM layer. Since this model is non-geometric, nodes are simply converted into rows in a feature vector, and no adjacency information is supplied.

In each trial, the root mean squared error (RMSE) was taken between the predicted and ground truth thicknesses for each of the 2003-2012 ice layers among the entire testing set. The mean RMSE and standard deviations were found for each year over all five trials. These results are showcased in Table \ref{table:OverallResults}. The proposed GAT-LSTM model performed significantly better than the baseline GCN and LSTM models both in terms of mean RMSE and consistency (shown by a lower standard deviation). The lack of adjacency data in the LSTM model and complex temporal learning in the GCN model may contribute to these results.

The overall performance of this model was numerically slightly poorer than what was displayed in our previous paper, though this is likely due to this model predicting more outputs (ten years' annual snow accumulation rather than only five) with less data (using only five years of historic snow accumulation data rather than ten) and use a larger and more diverse dataset.

\bibliographystyle{IEEEbib}
\bibliography{references}

\end{document}